\pdfoutput=1

\documentclass[11pt]{article}

\usepackage[preprint]{coling}

\usepackage{times}
\usepackage{latexsym}
\usepackage{booktabs,multirow}

\usepackage[T1]{fontenc}

\usepackage[utf8]{inputenc}

\usepackage{microtype}

\usepackage{inconsolata}

\usepackage{graphicx}

\title{WER We Stand: Benchmarking Urdu ASR Models}

\author{
 \textbf{Samee Arif\textsuperscript{1}},
 \textbf{Sualeha Farid\textsuperscript{2}},
 \textbf{Aamina Jamal Khan\textsuperscript{1}\textsuperscript{*}},
 \textbf{Mustafa Abbas\textsuperscript{1}\textsuperscript{*}},
\\
 \textbf{Agha Ali Raza\textsuperscript{1}},
 \textbf{Awais Athar\textsuperscript{3}},
\\
 \{samee.arif, 25100162, 25100079, agha.ali.raza\}@lums.edu.pk,\\
 sualeha@umich.edu, awais@ebi.ac.uk,
\\
 \textsuperscript{1}Lahore University of Management Sciences,
\\
 \textsuperscript{2}University of Michigan - Ann Arbor,
\\
 \textsuperscript{3}EMBL European Bioinformatics Institute
}

\begin{document}
\maketitle
\renewcommand{\thefootnote}{\fnsymbol{footnote}}
\footnotetext[1]{These authors contributed equally to this work.}
\renewcommand{\thefootnote}{\arabic{footnote}} 
\maketitle
\begin{abstract}
This paper presents a comprehensive evaluation of Urdu Automatic Speech Recognition (ASR) models. We analyze the performance of three ASR model families: \texttt{Whisper}, \texttt{MMS}, and \texttt{Seamless-M4T} using Word Error Rate (WER), along with a detailed examination of the most frequent wrong words and error types including insertions, deletions, and substitutions. Our analysis is conducted using two types of datasets, read speech and conversational speech. Notably, we present the first conversational speech dataset designed for benchmarking Urdu ASR models. We find that \texttt{seamless-large} outperforms other ASR models on the read speech dataset, while \texttt{whisper-large} performs best on the conversational speech dataset. Furthermore, this evaluation highlights the complexities of assessing ASR models for low-resource languages like Urdu using quantitative metrics alone and emphasizes the need for a robust Urdu text normalization system. Our findings contribute valuable insights for developing robust ASR systems for low-resource languages like Urdu.
\end{abstract}

\section{Introduction}
Automatic Speech Recognition (ASR) systems have become integral to modern technology, enabling numerous applications. Use cases include virtual assistants \citep{10.1007/978-981-99-9489-2_24} \citep{9210344}, smart homes \citep{Chen_2019} \citep{7990438}, medical assistance \citep{article} \citep{article1}, telecommunications \citep{659129}, and more. The ability to convert spoken language into text has revolutionized human-computer interaction, making technology more accessible and user-friendly.

ASR systems have made substantial progress in recent years, driven by advances in deep learning and the availability of large-scale datasets. However, the majority of this progress has been concentrated on resource-rich languages, leaving low-resource languages like Urdu with significant gaps in accuracy and reliability. Urdu, with over 70 million native speakers, is characterized by its rich phonetic diversity, complex morphological structure, and variety of regional dialects, all of which pose additional challenges for ASR systems. These challenges are further amplified in conversational contexts by informal speech patterns, code-switching \citep{khan2023codeswitchedurduasrnoisy}, and spontaneous speech disfluencies. The availability of annotated datasets is also limited compared to high-resource languages, which hinders the training and evaluation of ASR models.

Addressing these challenges is crucial for making ASR technology accessible and effective for Urdu speakers, particularly in scenarios involving both read and conversational speech. This paper focuses on evaluating and improving the performance of state-of-the-art ASR models and post-processing techniques for Urdu. Specifically, we examine three prominent ASR model families—\texttt{Whisper} \citep{radford2022robustspeechrecognitionlargescale}, \texttt{MMS} \citep{pratap2023scalingspeechtechnology1000}, and \texttt{Seamless-M4T} \citep{communication2023seamlessmultilingualexpressivestreaming} \citep{communication2023seamlessm4tmassivelymultilingual}—each of which has demonstrated strong performance across various languages.

In this paper, we introduce the first conversational speech dataset specifically designed for evaluating Urdu ASR models. We release all our fine-tuned models, datasets, evaluation scripts, and outputs to the community to foster further research in Urdu ASR tasks. By making these resources publicly available, we aim to bridge the gap in ASR technology for low-resource languages like Urdu, thereby contributing to more inclusive and effective speech recognition solutions.

Our contributions can be summarized as follows: 
\begin{enumerate} 
    \item We present the first conversational speech dataset for benchmarking Urdu ASR models. 
    \item We fine-tune ASR models from the \texttt{Whisper}, \texttt{MMS}, and \texttt{Seamless-M4T} families on Urdu, using both read speech and conversational speech datasets. 
    \item We conduct a comprehensive quantitative and qualitative analysis of non-fine-tuned and fine-tuned ASR models. We also highlight the complexity of evaluating Urdu ASR models using quantitative metrics alone, underscore the need for a robust text normalization system to address variations in word forms and improve overall model accuracy.
\end{enumerate}

The code, and model outputs are publicly available on GitHub\footnote{\url{https://github.com/sameearif/WER-We-Stand}}.

\section{Related Work}
\subsection{Modern ASR Models}
Modern ASR models, include \texttt{Wav2Vec2} \citep{baevski2020wav2vec20frameworkselfsupervised}, \texttt{Whisper}, \texttt{MMS} and \texttt{Seamless-M4T}. \texttt{Wav2Vec2}, introduced by Facebook AI, revolutionizes ASR by learning powerful speech representations from raw audio, enabling it to perform exceptionally well even with limited labeled data. \texttt{Whisper} by OpenAI, leverages large-scale datasets and transformers to achieve state-of-the-art accuracy across multiple languages. Similarly, Meta's \texttt{MMS} and \texttt{Seamless-M4T} models push the boundaries of multilingual and cross-modal ASR by incorporating vast amounts of diverse linguistic data. These advancements have significantly improved the ability of ASR systems to handle continuous, natural speech across a wide variety of languages and contexts.

\subsection{Low-Resource ASR}
The ASRoIL survey \citep{10.1007/s10462-019-09775-8} provides valuable insights for Indian languages into the challenges of variability in speech signals and the availability of corpora, detailing feature extraction techniques like MFCC, LPCC, and PLP, as well as various modeling and classification techniques. \citet{7955616}'s work offers an overview of ASR systems for South Indian languages, explaining methodologies and feature extraction methods by giving digests of seven papers. \citet{javed2022indicsuperbspeechprocessinguniversal} introduce the IndicSUPERB benchmark, featuring 1,684 hours of labeled speech data and establish benchmarks for six speech tasks. They train and evaluate different self-supervised models and demonstrate that language-specific fine-tuned models outperform baseline methods. \citet{app12178898}'s work systematically reviewed Arabic ASR studies from 2011 to 2021, covering feature extraction and classification techniques along with various aspects, such as the types of Arabic language supported, performance metrics, and existing research gaps. \citet{article2} focus on end-to-end deep learning frameworks, which integrate and simultaneously train the language model, pronunciation, and acoustic components, thus simplifying the pipeline. \citet{article3} presents a comparative study of Deep Neural Network (DNN) based Punjabi-ASR systems demonstrating that DNN-HMM hybrid models outperform traditional GMM-HMM architectures.

Low-resource languages have demonstrated significant advancements in multilingual settings, where models benefit from unexpected but advantageous learning capabilities. \citet{8461972} found that multilingual models exhibit cross-lingual transfer learning, which enhances the accuracy of low-resource languages by leveraging similarities and commonalities from high-resource languages. \citet{6639090} further analyzed cross-lingual transfer effects, finding that multilingual models fine-tuned on low-resource languages outperformed monolingual models, demonstrating the effectiveness of cross-lingual transfer. Overall, their work shows that low-resource languages like Urdu are more sensitive to model architecture due to data scarcity, and multilingual models can be trained effectively for both low-resource multilingual and monolingual tasks.

\subsection{Urdu ASR}
\citet{sharif2024statisticalmethodspretrainedmodels} provide an amalgamation of overviews of relevant studies on Urdu ASR, highlighting current trends, technological advancements, and future research directions, without directly comparing results on a common dataset. \citet{farooq19_interspeech} develop an Urdu LVCSR system using 300 hours of speech data and explores various acoustic modeling techniques, achieving a WER of 13.50\%,  though the models and data are not publicly available. \citet{5461790} developed a small-sized GMM using Word List Grammar by processing individual words and combining it with a Knowledge Base storing audio representations, pronunciations, and probabilities, achieving a Word Error Rate (WER) of 5.33 on seen speakers and 10.66 on unseen speakers. \citet{7571287} explored the impact of vocabulary size on HMM performance for Urdu ASR, finding an insignificant WER increase of 0.5\% between 100 and 250-word vocabularies. \citet{9333026} used a GMM combined with a graphene-to-phone LSTM CMUsphinx and KENLM to calculate and store n-gram probabilities, achieving a 9.64\% WER, significantly improving accuracy. \citet{khan2023codeswitchedurduasrnoisy} combined HMM and CNN-TDNN with LF-MMI, achieving a 13\% WER.

\section{Models and Datasets}
\subsection{ASR and Post-Processing Models}
In this paper we evaluate 9 ASR models given in Table \ref{tab:asr-models}.

\begin{table}[h]
\centering
\small
\begin{tabular}{ll}
\toprule
\textbf{Family} & \textbf{Model} \\ \midrule
Whisper & Tiny, Base, Small, Medium, \\
                 & Large-v3 \\[.5em]
MMS & 300 Million, 1 Billion \\[.5em]
Seamless-M4T & Medium, v2-Large \\
\bottomrule
\end{tabular}
\caption{ASR model evaluated in this paper.}
\label{tab:asr-models}
\end{table}

\subsection{ASR Datasets}
We evaluate the ASR models on two types of datasets: (1) Read Speech, and (2) Conversational Speech. We developed and present the conversational speech dataset in this paper.

\subsubsection{Read Speech}
We use the ARL Urdu Speech Database\footnote{\url{https://catalog.ldc.upenn.edu/LDC2007S03}} which has 159,996 audios. The distribution of speaker dialects in the corpus is given in Table \ref{tab:accent-distribution}.

\begin{table}[h!]
\centering
\small
\begin{tabular}{|l|c|}
    \hline
    \rule{0pt}{2.5ex}\textbf{Accent} & \textbf{Number of Speakers} \\ \hline
    \rule{0pt}{2.5ex}South Sindh & 29 \\ \hline
    \rule{0pt}{2.5ex}North Sindh & 30 \\ \hline
    \rule{0pt}{2.5ex}South Punjab & 27 \\ \hline
    \rule{0pt}{2.5ex}North Punjab & 29 \\ \hline
    \rule{0pt}{2.5ex}Capital Area & 29 \\ \hline
    \rule{0pt}{2.5ex}North West Regions & 30 \\ \hline
    \rule{0pt}{2.5ex}Baluchistan & 26 \\ \hline
\end{tabular}
\caption{Distribution of Speakers Across Different Accents in the Dataset}
\label{tab:accent-distribution}
\end{table}

For the dataset creation each speaker is presented with 400 prompts to read: sentences, place names, and person names. Two microphones set at different distances to the speaker are used for the recordings. Punctuation are omitted and numbers were written out in full.

We also evaluate the models using the bona fide audio files from the CSaLT Deepfake dataset \citep{munir-etal-2024-deepfake}, which includes 6 female and 11 male speakers, with a total of 3,398 audio files amounting to 42.9 hours of recordings.

\subsubsection{Conversational Speech}
We present the first Urdu conversational speech dataset, which consists of 471 audio recordings (1.3 hours). These audios were recorded through calls over the internet. This is to mirror actual conversational environments and meetings, making the data more relevant and practical for real-world applications.

The dataset features 4 female and 6 male speakers who used the microphones they had on hand to replicate real-life audio quality. To ensure smooth and natural conversations, the participants, all native Urdu speakers and computer science students, were asked to form groups and pairs with people they felt comfortable with. The recordings were done in various group sizes. Four sessions involved groups of 2, One session involved a group of 3, and another session involved a group of 4.

Participants were also encouraged to choose topics they were comfortable discussing, which resulted in a diverse range of discussions. The final topics included Pakistan Independence Day, Group Projects, Ramadan and Eid, Neighbors Discussing Load-shedding and Prices, Health Issues, and Gossips.

For transcription, the process was carried out in three passes to ensure accuracy and consistency. The first pass transcription was completed by the original recorders themselves. This was followed by a second pass, where two research interns, who were not involved in the recording, refined the transcriptions and split the audio into smaller chunks. The final pass was done by two of the authors of this paper, ensuring the highest level of accuracy in the transcriptions.

\section{Experimental Design}
We fine-tune the ASR models in Table \ref{tab:asr-models} using a combined dataset of Mozilla's Common Voice \citep{commonvoice:2020} and 
Google's Fleurs \cite{conneau2022fleursfewshotlearningevaluation}, which together provide 16,156 audio samples. Given the limited amount of training data available, we opted for a 90-10 split, with 90\% of the data used for training and the remaining 10\% for validation. We evaluate both the non-fine-tuned and the fine-tuned models on ARL and CSaLT dataset for read speech and on our dataset for conversational speech. \texttt{mms-300m} base model is not fine-tuned for downstream task of speech recognition so we only evaluate our fine-tuned version. Before conducting the evaluations, we normalize both the ground truth and the model predictions. This involves removing punctuation and disfluencies. This preprocessing step ensures that the evaluations focus on meaningful transcription accuracy rather than being skewed by minor formatting inconsistencies or speech disfluencies and punctuation marks. We noticed that Whisper and MMS models have built-in disfluency filtering and perform this task effectively. However, Seamless models output disfluencies marked with a \# (e.g., \#um), which we removed using regex during preprocessing.

\section{Results and Discussion}
Table \ref{tab:asr-results} shows the WER of the ASR models, fine-tuned and non-fine-tuned, on read speech datasets—ARL and CSaLT—and our conversational speech dataset.

\begin{center}
\begin{table*}[htbp]
\centering
\renewcommand{\arraystretch}{1}
\small 
\begin{tabular}{lcccccccccccc}
\toprule
& \multicolumn{2}{c}{\textbf{Read Speech (ARL)}} & \multicolumn{2}{c}{\textbf{Read Speech (CSaLT)}} & \multicolumn{2}{c}{\textbf{Conversational Speech}} \\
\cmidrule(lr){2-3} \cmidrule(lr){4-5} \cmidrule(lr){6-7}
& \textbf{Base Model} & \textbf{Fine-tuned} & \textbf{Base Model} & \textbf{Fine-tuned} & \textbf{Base Model} & \textbf{Fine-tuned} \\
\textbf{Model} \\
\midrule
\texttt{whisper-tiny} & 116.92 & 45.59 & 96.57 & 42.12 & 163.18 & 59.99 \\
\texttt{whisper-base} & 71.53 & 39.84 & 57.77 & 38.86 & 163.52 & 48.61 \\
\texttt{whisper-small} & 48.70 & 28.60 & 41.10 & 27.39 & 55.67 & 32.92 \\
\texttt{whisper-medium} & 37.04 & 25.38 & 33.39 & 24.15 & 40.22 & 28.87 \\
\texttt{whisper-large} & 26.25 & 23.79 & 24.44 & 22.35 & \textbf{18.30} & \textbf{17.86}\\
\texttt{mms-300m} & - & 51.48 & - & 47.73 & - & 66.40 \\
\texttt{mms-1b} & 39.65 & 28.37 & 34.60 & 26.85 & 46.44 & 42.46 \\
\texttt{seamless-medium} & 30.06 & 19.41 & 24.18 & 20.59 & 22.33 & 20.01 \\
\texttt{seamless-large} & \textbf{23.97} & \textbf{17.09} & \textbf{20.57} & \textbf{18.61} & 29.99 & 18.75 \\
\bottomrule
\end{tabular}
\caption{WER of ASR models on ARL and CSaLT dataset for Read Speech and on our dataset for Conversational Speech. For each category, bold represents the lower WER in the specified category.}
\label{tab:asr-results}
\end{table*}
\vskip -0.2in
\end{center}

\subsection{Read Speech}
Read speech typically consists of well-articulated sentences with fewer disfluencies, making it a less complex task for ASR models compared to conversational speech. However, challenges such as dialectal variations and pronunciation differences still impact transcription accuracy.

\subsubsection{Quantitative Analysis}
\begin{figure}[h!]
    \begin{center}
    \centerline{\includegraphics[width=\columnwidth]{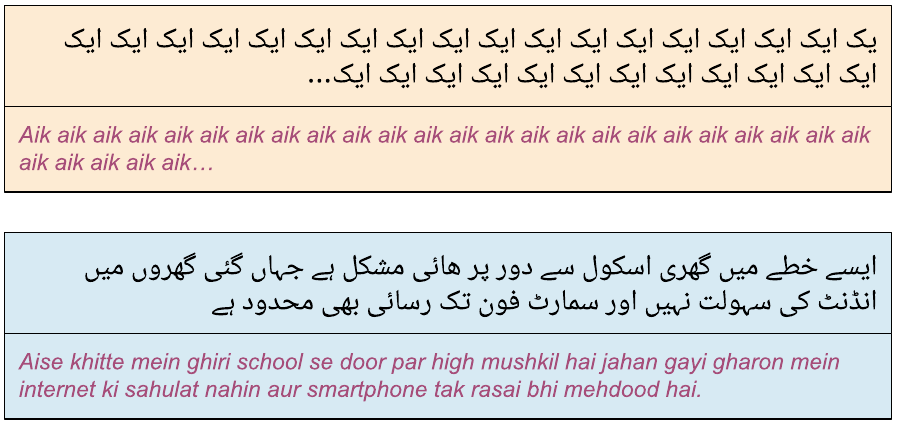}}
    \caption{\small{The yellow and blue boxes represent the output of the same audio on general and  fine-tuned  \texttt{whisper-tiny} models respectively}}
    \label{fig:tiny-hallucinating}
    \end{center}
    \vskip -0.3in
\end{figure}

In both the ARL and CSaLT datasets, the evaluated ASR models reveal distinct patterns in performance based on their architecture and size. Smaller models, such as \texttt{whisper-tiny} and \texttt{whisper-base}, suffer from significant hallucination issues in their base versions, as indicated by the high WERs—116.92 and 96.57 on ARL for \texttt{whisper-tiny} and \texttt{whisper-base}, respectively. Similarly, on CSaLT, the WER for \texttt{whisper-tiny} is 96.57, indicating that these models struggle to handle Urdu without fine-tuning. This issue is depicted in Figure \ref{fig:tiny-hallucinating}, where the non-fine-tuned \texttt{whisper-tiny} generates incorrect transcriptions. However, fine-tuning substantially reduces these error rates, as seen with \texttt{whisper-tiny}'s drop to 45.59 on ARL and 42.12 on CSaLT, and \texttt{whisper-base}'s improvement to 39.84 and 38.86 on the respective datasets. Despite these improvements, the smaller models remain limited due to their relatively small parameter sizes—34M for \texttt{whisper-tiny} and 74M for \texttt{whisper-base}.

Moving to larger models, \texttt{whisper-small} (244M parameters) and \texttt{whisper-medium} (769M parameters) demonstrate a notable improvement over their smaller counterparts. For instance, \texttt{whisper-small} reduces its WER from 48.70 to 28.60 on ARL after fine-tuning, while \texttt{whisper-medium} improves from 37.04 to 25.38. The trend is consistent on CSaLT, where \texttt{whisper-small} drops from a WER of 41.10 to 27.39 and \texttt{whisper-medium} from 33.39 to 24.15. However, the largest model, \texttt{whisper-large} with 1.55B parameters, performs better, achieving a WER of 23.79 on ARL and 22.35 on CSaLT after fine-tuning. Although \texttt{whisper-large} shows impressive results, the \texttt{seamless-m4t} models outperform it. For instance, \texttt{seamless-medium} with 1.2B parameters achieves a WER of 30.06 pre-fine-tuning and 19.41 post-fine-tuning on ARL. Additionally, fine-tuned \texttt{seamless-large} consistently outperforms all other models, with a WER of 17.09 on ARL and 18.61 on CSaLT, showcasing the efficiency of the Seamless-M4T architecture.

In contrast, the \texttt{mms} models, despite their large parameter sizes, underperform. The \texttt{mms-300m} model struggles, with a WER of 51.48 on ARL and 47.73 on CSaLT datasets, whereas \texttt{mms-1b} achieves more competitive results after fine-tuning, reducing its WER from 39.65 to 28.37 on ARL and from 34.60 to 26.85 on CSaLT. \texttt{mms-300m} has a lower WER on both datasets compared to \texttt{whisper-tiny} with 39M parameters. The fine-tuned \texttt{mms-1b} has comparable performance to \texttt{whisper-small} with 244M parameters. This indicates the MMS family is outperformed by much smaller models from other model families, highlighting that model size alone does not guarantee better performance, and other factors such as architecture and training data play a crucial role.

\begin{figure}[h!]
    \begin{center}
    \centerline{\includegraphics[width=\columnwidth]{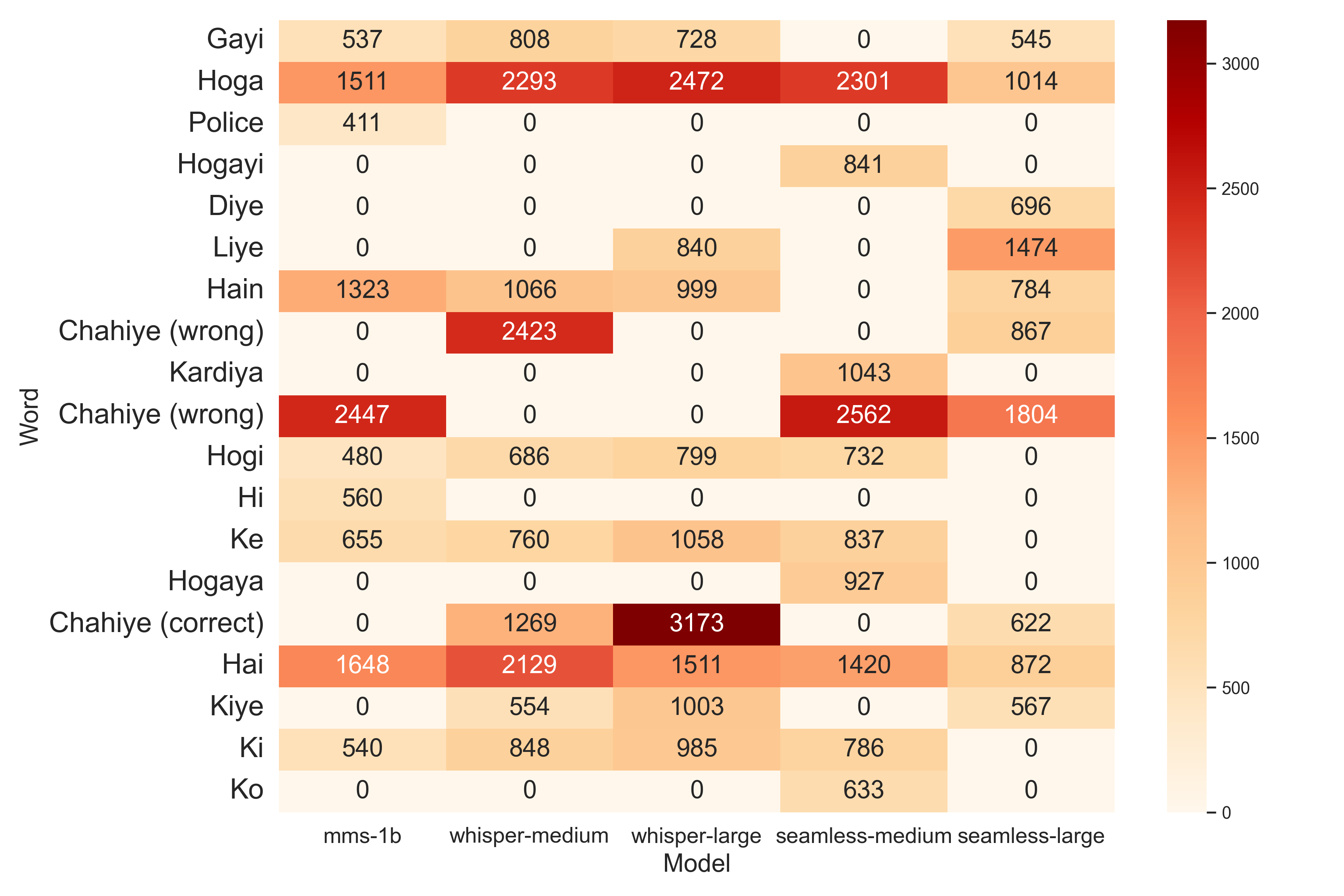}}
    \caption{\small{Comparison of wrong words across non-fine-tuned models.}}
    \label{fig:heatmap_comparison-nft}
    \end{center}
    \vskip -0.2in
\end{figure}

The heatmap in Figure \ref{fig:heatmap_comparison-nft} compares the most frequent errors (wrong words) on combined ARL and CSaLT datasets across five non-fine-tuned models: \texttt{mms-1b}, \texttt{whisper-medium}, \texttt{whisper-large}, \texttt{seamless-medium}, and \texttt{seamless-large}.

Several words appear to stand out as high-frequency errors across all models. Notably, "Ho ga" and "Chahiye" are consistently problematic, with particularly high error counts in \texttt{whisper-medium}, \texttt{whisper-large}, and \texttt{seamless-large}. The word "Chahiye" appears with orthographic variations in Urdu—one with two "Yeh" and no "Hamza" character, one with one "Yeh" and a "Hamza," and one with with no "Yeh" and a "Hamza." The first variation is the most widely accepted correct form, though all are used interchangeably in practice. These errors can likely be attributed to the training data, which often includes user-generated content like internet text, where such variations are common.

"Chahiye" has an error frequency of 1,269 in \texttt{whisper-medium} and 3,173 in \texttt{whisper-large}, indicating that these models do not reproduce the  specific spelling of this word as used in the dataset. \texttt{mms-1b} shows relatively fewer errors for many words in this specific analysis, but it has a higher WER for the ARL and CSaLT datasets. This indicates that, while fewer individual words may be highlighted as problematic here, the model struggles with a broader range of words overall, leading to a higher overall error rate. The word "Hoga" shows a high error rate across all models, as it is reproduced as "Ho ga" in the model transcriptions, but the test dataset contains it as a single word, "Hoga." This mismatch leads to consistent false positives, where the models’ correct transcription is flagged as incorrect due to the way the word is represented in the test dataset. The Whisper family has the higher tendency to misrecognize the word "Hai" followed by MMS and then Seamless.

\begin{figure}[h!]
    \begin{center}
    \centerline{\includegraphics[width=\columnwidth]{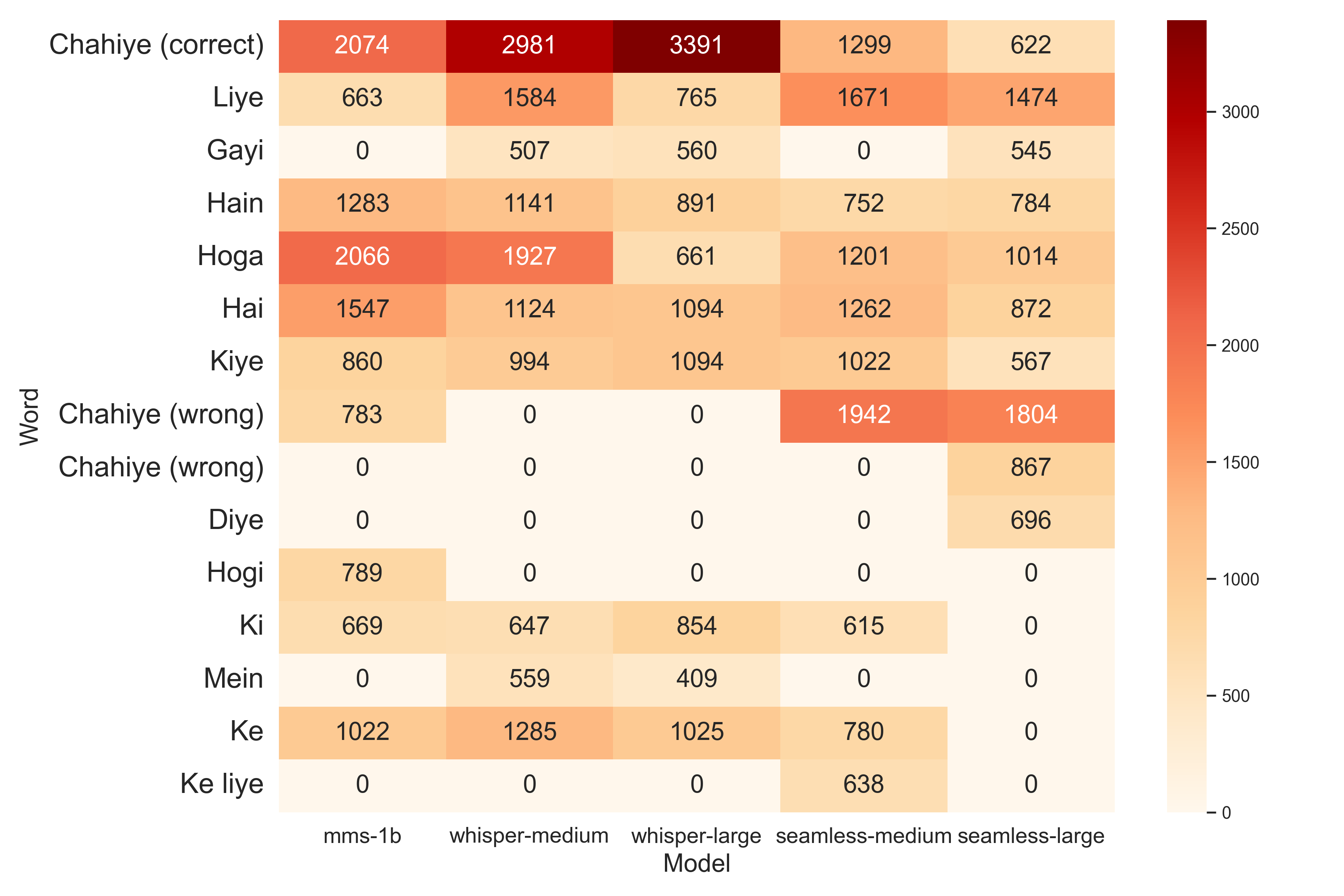}}
    \caption{\small{Wrong words across fine-tuned models.}}
    \label{fig:heatmap_comparison-ft}
    \end{center}
    \vskip -0.3in
\end{figure}

Figure \ref{fig:heatmap_comparison-ft} compares the most frequent errors on ARL and CSaLT datasets across the fine-tuned models. In the fine-tuned models, certain high-frequency errors observed in the non-fine-tuned models have been reduced, indicating that fine-tuning has helped address some transcription challenges. However in some cases the error rate has gone up. For instance, the error rate for the word "Chahiye" (expected spellings) has increased after fine-tuning. We found 155 misspelled versions of this word in Common Voice and Fleurs dataset. For \texttt{seamless-medium}, the error has gone up from 0 to 1942. \texttt{whisper-large} still shows a high error with slight increase from 3173 to 3391. The word "Ho ga" also poses a challenge for fine-tuned models. There are 82 instances of "Ho ga" and 87 instances of "Hoga" in the training dataset. As a result, the false positives seen earlier are now turned into inconsistencies, where the models are confused between the two variations, leading to unpredictable outputs. A lattice of plausible respellings of the reference transcription, as done so by \citealp{karita-etal-2023-lenient} may help address this issue. By employing a lenient evaluation with orthographical variations in mind, these false errors could be overlooked.

The words like "Hi", "Ki" and "Ko" show improved performance in the fine-tuned models. While fine-tuning has led to notable improvements in transcription accuracy for several words, persistent challenges remain, particularly for high-frequency words and segmentation issues. This evaluation highlights the complexity of evaluating Urdu ASR models using quantitative metrics alone and underscores the need for a robust text normalization system to address variations in word forms and improve overall model accuracy.

\subsubsection{Qualitative Analysis}
Fine-tuning leads to substantial improvements across all models, the \texttt{seamless-large} emerges as the best performer across the ARL and CSaLT datasets. The performance of the \texttt{seamless-medium} and \texttt{seamless-large} models can be attributed to several key differences in training data and architecture. Seamless-M4T benefited from over 470,000 hours of multimodal speech and text data, sourced from the SeamlessAlign dataset, which was specifically crafted to support a broad spectrum of languages. This extensive, diverse dataset, combined with the use of SONAR embeddings—designed to offer modality and language-agnostic representations—enhances Seamless-M4T’s ability to generalize effectively across various languages and speech domains. The architecture supports more complex multimodal learning, contributing to its edge in performance, particularly in tasks involving low-resource languages like Urdu.

In contrast, \texttt{whisper-large} was trained on approximately 1 million hours of labeled data and 4 million hours of pseudo-labeled data, with a focus on automatic speech recognition (ASR) tasks. While Whisper is capable of handling numerous languages, its emphasis is primarily on transcription accuracy. Without the same multimodal and speech-to-speech training exposure found in Seamless-M4T, Whisper lacks the cross-modal adaptability that contributes to the superior results seen in Seamless-M4T’s performance on read speech tasks.

Similarly, while the MMS models were pretrained on 500,000 hours of speech data across 1,400 languages, their focus on Wav2Vec2-based self-supervised learning may explain their relative underperformance. MMS models are designed to excel at learning robust speech representations without labeled data, but they lack the comprehensive multimodal and multitask learning present in Seamless-M4T. This limits their performance on specific tasks like Urdu read speech, where cross-modal training and fine-tuning play a crucial role. Thus, despite extensive pretraining, MMS models are unable to match the fine-tuned, multimodal capabilities of Seamless-M4T in such applications.

\subsection{Conversational Speech}
The results for conversational speech highlight distinct performance patterns among ASR models when compared to read speech, mainly due to the inherent challenges posed by spontaneous, natural dialogue. Conversational speech often includes disfluencies, overlaps, and variations in speaker styles, making it more complex to transcribe accurately.

\subsubsection{Quantitative Analysis}
Most models show higher WERs in conversational speech, reflecting the increased difficulty of this task. Among the Whisper models, \texttt{whisper-large} stands out with the lowest WER of 18.30 in its base version, slightly improving to 17.86 after fine-tuning. This positions it as the best-performing model for conversational speech. In contrast, smaller models like \texttt{whisper-tiny} and \texttt{whisper-base} struggle significantly. Their base versions exhibit extremely high WERs—163.18 and 163.52, respectively—due to hallucinations and transcription errors. Fine-tuning reduces these errors, bringing their WERs down to 59.99 and 48.61, but they remain far behind the larger models. Mid-range models, such as \texttt{whisper-small} (244M parameters) and \texttt{whisper-medium} (769M parameters), also improve with fine-tuning, achieving WERs of 32.92 and 28.87, respectively.

The MMS models, despite their large parameter sizes, underperform in conversational speech. The \texttt{mms-1b} model, shows some improvement after fine-tuning, reducing its WER from 46.44 to 42.46, but it still lags behind even the smaller Whisper models like \texttt{whisper-medium}. The \texttt{mms-300m} model performs worse, with a post-fine-tuning WER of 66.40, suggesting that the MMS models' pretraining data and objectives do not generalize well to conversational speech in low-resource languages such as Urdu.

The Seamless-M4T models also perform well on conversational speech, though their strengths lie primarily in read speech tasks. Notably, the smaller \texttt{seamless-medium} model initially outperforms its larger counterpart, \texttt{seamless-large}, in its base version, achieving a WER of 22.33 compared to 29.99 for the large model. This suggests that the medium model, despite having fewer parameters, handles the nuances of conversational speech better. However, after fine-tuning, \texttt{seamless-large} shows more significant improvement, reducing its WER to 18.75, while \texttt{seamless-medium} only improves to 20.01. This indicates that while the medium model may have an edge in its base state, the larger model benefits more from fine-tuning, ultimately surpassing the performance of the medium model.

Overall, the results for conversational speech show that larger Whisper models and Seamless-M4T are better equipped to handle the complexities of spontaneous dialogue. Whisper models, particularly \texttt{whisper-large}, exhibit a slight edge over Seamless-M4T, although the gap is narrower compared to their performance in read speech tasks. The MMS models, despite their large size, do not perform as well in this task, highlighting potential limitations in their training objectives. Fine-tuning proves to be essential for improving performance across all models when dealing with conversational speech.

\begin{figure}[h!]
    \begin{center}
    \centerline{\includegraphics[width=\columnwidth]{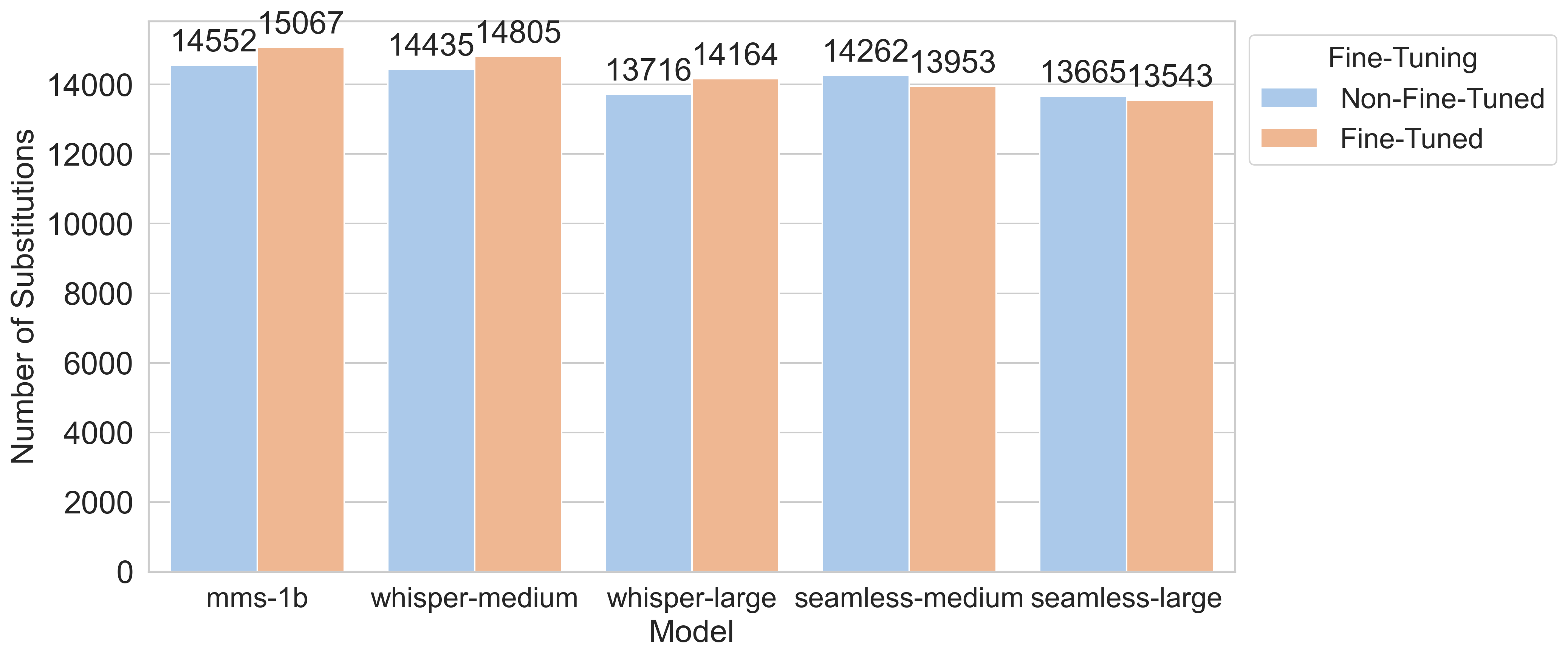}}
    \caption{\small{Substitution errors across models before and after fine-tuning.}}
    \label{fig:substitutions_per_model_joint}
    \end{center}
    \vskip -0.2in
\end{figure}

\begin{figure}[h!]
    \begin{center}
    \centerline{\includegraphics[width=\columnwidth]{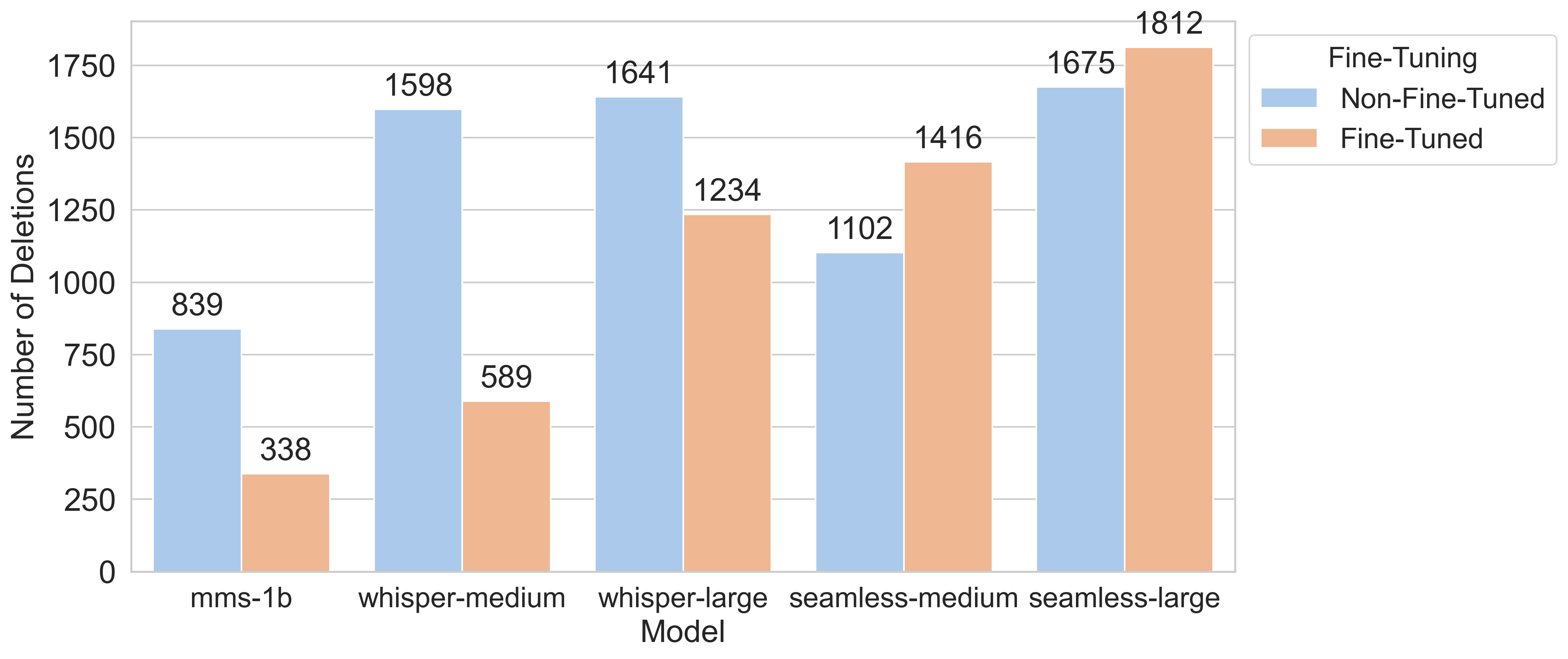}}
    \caption{\small{Deletion errors across models before and after fine-tuning.}}
    \label{fig:deletion_per_model_joint}
    \end{center}
    \vskip -0.2in
\end{figure}

The analysis of the conversational dataset on non-fine-tuned models reveals significant variability in error patterns across the different ASR models. While all models show high substitution errors, there are notable differences in the number of deletions and insertions. Figure \ref{fig:substitutions_per_model_joint} gives the comparison of substitution errors and Figure \ref{fig:deletion_per_model_joint} gives the comparison of deletion errors across models before and after fine-tuning. Substitutions dominate the total errors for all models, with counts ranging from 13,665 for \texttt{seamless-large} to 14,552 for \texttt{mms-1b}, indicating that non-fine-tuned models frequently mis-recognize words, leading to incorrect transcriptions. This is especially problematic for conversational datasets where context and word prediction are key factors for accurate transcription.

Deletions also account for a substantial portion of errors, particularly for \texttt{whisper-large} (1,641) and \texttt{whisper-medium} (1,598). High deletion errors suggest that these models are often failing to transcribe words altogether, which can significantly distort the meaning of the conversation. Interestingly, \texttt{mms-1b} shows the lowest number of deletions (844), indicating that while it may substitute many words, it tends to capture more of the spoken content compared to other models. Insertions remain minimal across all models (non-fine-tuned and fine-tuned), with each registering only a few insertion errors (1 to 4). This suggests that these models rarely add extra words that were not spoken.

After fine-tuning, the overall performance of the ASR models shows notable changes, though the error patterns remain similar in terms of overall trends. Substitution errors continue to account for the majority of the total errors. We can observe in Figure \ref{fig:substitutions_per_model_joint} and \ref{fig:deletion_per_model_joint} that after fine-tuning the WER of Whisper models and MMS goes down because of reduction in deletion errors and the WER of Seamless models go down because of the reduction in substitution errors. Fine-tuning seems to reduce substitution errors for \texttt{seamless-large} to 13,543 and \texttt{seamless-medium} to 13,953 but it increased slightly for other models. Deletions error particularly, for \texttt{mms-1b} saw a drop from 844 to 338, and for \texttt{whisper-medium} dropped from 1,598 to 589. This indicates that fine-tuning helps the models transcribe more spoken content in their transcriptions.

\subsubsection{Qualitative Analysis}
\begin{figure}[h!]
    \begin{center}
    \centerline{\includegraphics[width=\columnwidth]{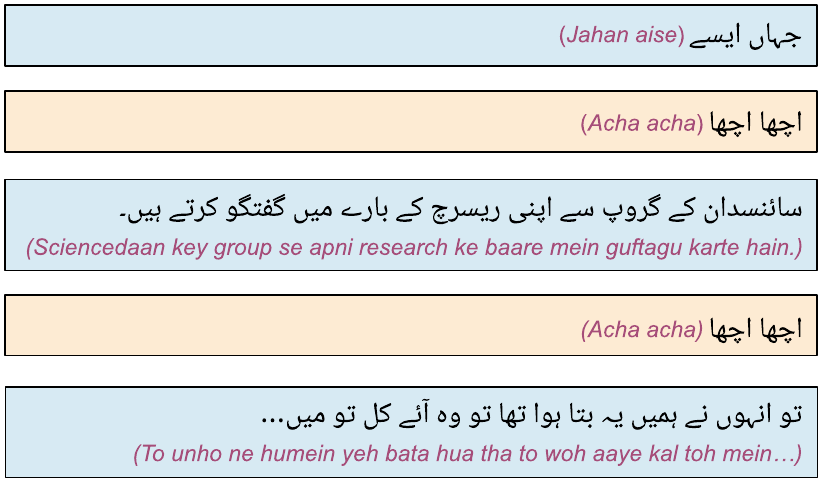}}
    \caption{\small{The blue color represents speaker 1 and yellow represents speaker 2.}}
    \label{fig:overlapping-speakers}
    \end{center}
    \vskip -0.3in
\end{figure}

A significant limitation across all ASR models is their inability to handle overlapping speech from multiple speakers. Figure \ref{fig:overlapping-speakers} illustrates an example of ground truth for overlapping speech, where the audio from speaker 1 and speaker 2 overlaps. As shown in the figure, speaker 1's audio is dominant, causing ASR models to fail in accurately transcribing the speech of speaker 2.

\begin{figure}[h!]
    \begin{center}
    \centerline{\includegraphics[width=\columnwidth]{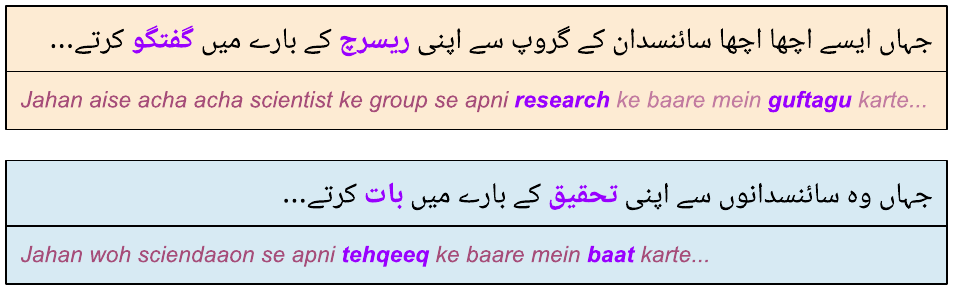}}
    \caption{\small{The yellow box represents the ground truth and the blue ones represents the prediction.}}
    \label{fig:seamless-errors}
    \end{center}
    \vskip -0.3in
\end{figure}

The higher WER of base \texttt{seamless-large} can be partly attributed to two distinct factors: its handling of English words within the Urdu conversational dataset and its tendency to paraphrase certain words. Instead of transliterating English words, it often translates them, leading to errors. For example, as shown in Figure \ref{fig:seamless-errors}, the word "Research" is translated to "Tehqeeq" rather than being transliterated. Additionally, the model tends to paraphrase words within the Urdu content, such as transcribing "Guftagu" as "Baat". These errors likely stem from the model's training on a multimodal dataset that emphasizes cross-language translation and broader contextual understanding, rather than strict transcription accuracy. Seamless-M4T's architecture, designed for tasks beyond ASR—such as speech-to-speech translation—encourages this behavior. The model prioritizes conveying meaning across languages, which leads to translations and paraphrasing when encountering multilingual inputs, as opposed to Whisper, which is primarily trained for transcription fidelity.

\section{Conclusion}
In this paper, we present a comprehensive evaluation of three ASR model families—Whisper, MMS, and Seamless-M4T—on Urdu read and conversational speech datasets. Our findings highlight the challenges associated with developing robust ASR systems for low-resource languages like Urdu, particularly when faced with spontaneous conversational speech, disfluencies, and code-switching. Among the models, \texttt{whisper-large} and \texttt{seamless-large} stand out, with \texttt{whisper-large} excelling in conversational contexts and \texttt{seamless-large} demonstrating strong performance in read speech tasks. Despite improvements from fine-tuning, challenges such as handling overlapping speech and distinguishing between similar phonetic patterns remain prevalent across models, indicating the need for further refinement. Additionally, our error analysis reveals the importance of text normalization and highlights the potential of multimodal approaches to improve ASR accuracy. This study contributes valuable insights into the capabilities and limitations of current ASR models for Urdu and underscores the importance of designing specialized datasets and evaluation metrics. Our work will also be valuable for developers looking to build real-world applications, such as virtual assistants, voice-controlled devices, and transcription services.

\section{Future Work}
This study opens several avenues for future exploration in Urdu ASR. Addressing overlapping speech in conversational settings remains a critical challenge. Future work could focus on evaluating speaker diarization using our dataset to improve multi-speaker recognition. Enhancing the handling of code-switching between Urdu and English, especially for models like \texttt{seamless-large} that tend to translate rather than transliterate, could significantly boost transcription accuracy in real-world contexts where multilingualism is common. Exploring the potential of Large Language Models (LLMs), such as \texttt{GPT-4o} and \texttt{Llama-3.1}, for ASR output post-processing could further refine transcription quality by correcting errors and improving fluency. Finally, the integration of a lenient evaluation method which incorporates orthographical variations, or robust text normalization system tailored for Urdu could address variations in spelling and word forms, improving the consistency of transcriptions. Extending this work to cross-modal tasks, such as speech-to-speech translation or text generation, could provide valuable insights for building more advanced systems that handle complex, multimodal language tasks for low-resource languages like Urdu.

\section*{Acknowledgments}
We are deeply grateful to Sualeha Farid (University of Michigan - Ann Arbor) for her invaluable contributions, including conducting data generation sessions and assisting with data analysis, without which this paper would not have been possible. 

We are also grateful to our research interns: Muhammad Abubakar Mughal, Natiq Khan, Anum Javed, Aleena Saqib, Muhammad Abdullah Sohail, Bushra Zubair, Hamza Iqbal, Salaar Masood, Maham Javed, Faisal Haider, Muhammad Suhaib Rashid, and Muhammad Faizan Waris. Their dedication and hard work in creating and refining the dataset were instrumental to the success of this project. We sincerely appreciate their invaluable contributions and commitment to our research efforts.

\bibliography{custom}

\end{document}